\documentclass[sigconf,screen,preprint]{acmart}

\usepackage{amssymb}
\usepackage{subcaption}
\usepackage{caption}
\usepackage{graphicx}
\usepackage{todonotes}
\usepackage{booktabs}

\setcopyright{none}
\renewcommand\footnotetextcopyrightpermission[1]{}

\acmConference[UrbComp'17]{ACM KDD conference}{August 14, 2017}{Halifax, Nova Scotia, Canada} 
\acmYear{2017}
\acmDOI{}
\acmISBN{}
\acmPrice{}


\makeatletter
\def\runningfoot{\def\@runningfoot{ab}}
\def\firstfoot{\def\@firstfoot{cd}}
\makeatother

\title{Automated Lane Detection in Crowds using Proximity Graphs}
\date{\today}

\author{Stijn Heldens}
\email{s.j.heldens@utwente.nl}
\affiliation{
  \institution{University of Twente}
  \country{the Netherlands}
}

\author{Claudio Martella}
\email{claudio.martella@vu.nl}
\affiliation{
  \institution{VU University Amsterdam}
  \country{the Netherlands}
}

\author{Nelly Litvak}
\email{n.litvak@utwente.nl}
\affiliation{
  \institution{University of Twente}
  \country{the Netherlands}
}

\author{Maarten van Steen}
\email{m.r.vansteen@utwente.nl}
\affiliation{
  \institution{University of Twente}
  \country{the Netherlands}
}


\newcommand{\reffig}[1]{Figure \ref{fig:#1}}
\newcommand{\refsec}[1]{Section \ref{sec:#1}}
\newcommand{\ignore}[1]{}

\begin{document}
\begin{abstract}
Studying the behavior of crowds is vital for understanding and predicting human interactions in public areas. Research has shown that, under certain conditions, large groups of people can form \emph{collective behavior patterns}: local interactions between individuals results in global movements patterns. To detect these patterns in a crowd, we assume each person is carrying an on-body device that acts a local proximity sensor, e.g., smartphone or bluetooth badge, and represent the \emph{texture} of the crowd as a \emph{proximity graph}. Our goal is extract information about crowds from these proximity graphs. In this work, we focus on one particular type of pattern: lane formation. We present a formal definition of a lane, proposed a simple probabilistic model that simulates lanes moving through a stationary crowd, and present an automated lane-detection method. Our preliminary results show that our method is able to detect lanes of different shapes and sizes. We see our work as an initial step towards rich pattern recognition using proximity graphs.
\end{abstract}

\maketitle
\renewcommand{\shortauthors}{S. Heldens et. al.}

\let\thefootnote\relax\footnotetext{\large Copyright is held by the author/owner(s). \\
UrbComp'17, August 14, 2017, Halifax, Nova Scotia, Canada.}
\section{Introduction}
\label{sec:introduction}

\begin{figure}
  \centering
  \includegraphics[width=0.5\textwidth]{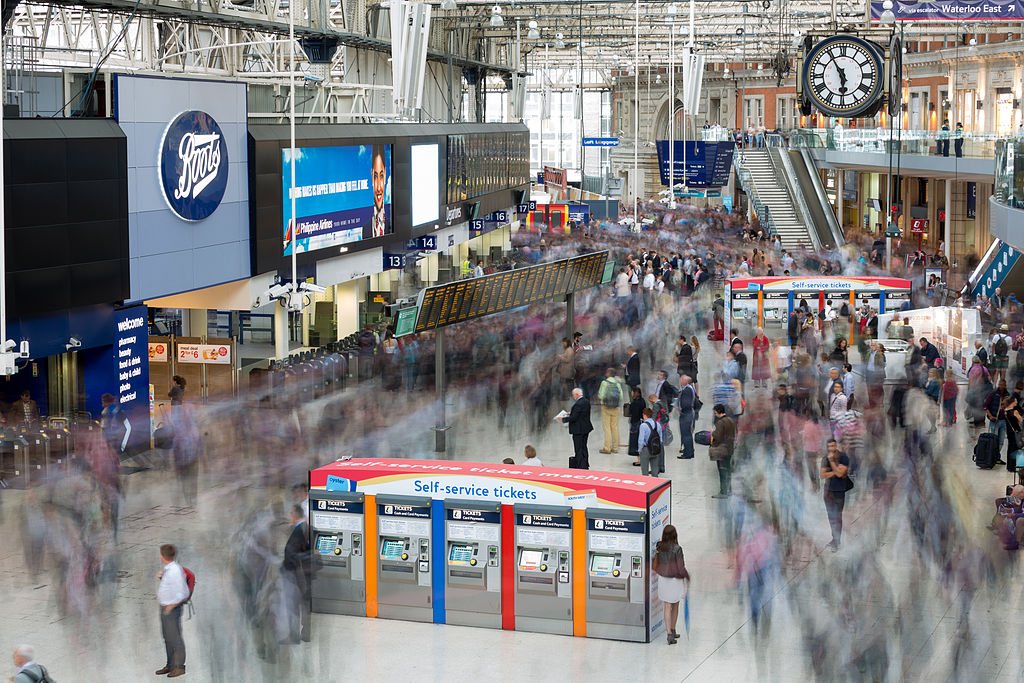}
  \caption{Long exposure shot at busy train station reveals lane formation. Photo by David Iliff, CC-BY-SA 3.0 license.}
  \label{fig:example_of_real_lane}
\end{figure}

Different studies (see survey by Castellano et al.~\cite{castellano2009statistical}) have shown that, while the behavior of individuals in public areas is often erratic and unpredictable, the behavior of large crowds as a whole is predictable and can be modeled. Crowd simulation models are plentiful, examples are models based on fluid dynamics~\cite{helbing1998fluid,guo2008mobile}, cellular automata~\cite{sarmady2011cellular}, or dynamic systems~\cite{helbing1995social}.


Helbing et al. observed that crowds have a tendency to form \emph{collective movement patterns}~\cite{helbing2001self}. The patterns are not globally planned or externally organized, but emerge naturally from the local interactions between individuals. Examples are circulation of flow at intersections, clogging at bottlenecks, and formation of lanes in crowded areas. Automated detection of these patterns is crucial for understanding, analyzing, and predicting the behavior of crowds in large open areas. One can think of a large number of applications~\cite{zhan2008crowd}, for example, improve safety at sport matches, music concerts, or public demonstrations, provide guidelines for urban planners to improve design of public spaces, or automate detection of anomalies.

Previous attempts at automated detection of these patterns utilize surveillance cameras combined with image processing techniques (see references in \refsec{related_work}). These techniques are, however, inherently limited to the perspective of one camera. In our work, instead of employing cameras, we assume each person is wearing a device that acts as a local proximity sensor: each sensor can detect other sensors nearby. These devices can be implemented using readily available hardware such as smartphones or electronic badges~\cite{martella2014proximity}. 

Each detection between two devices corresponds to an edge in a graph which changes over time, a so-called \emph{proximity graph}~\cite{martella2014crowd}. A proximity graphs characterizes the \emph{texture} of a crowd and describes how individuals navigate through the space. While a single camera can only cover a small area, proximity graphs provide a holistic view of large areas.

Extracting global movement patterns from proximity graphs is challenging since each device provides only local information. The fundamental problem that we tackle is how to uncover global patterns based on local detections.

In this work, we focus on one particular type of pattern: the formation of \emph{lanes}. Lanes often appear in crowds when groups of people traverse a densely crowded space, for example, in a narrow shopping street or at a busy train station (\reffig{example_of_real_lane}). We propose an automated lane-detection method based on proximity graphs. Our method combines techniques from graph embedding with a density-based clustering algorithm to identify lanes. To evaluate our method, we present a model that simulates lanes moving through a stationary crowd. Preliminary results show that our method is able to detect lanes of different sizes and shapes. Overall, our work can be seen as the first step towards rich motion pattern recognition using proximity graphs. 

The remaining sections are structured as follows: \refsec{background} presents background information, \refsec{algorithm} describes our lane-detection method, \refsec{model} proposes the simulation model, \refsec{evaluation} \& \ref{sec:results_sec} present results, \refsec{related_work} describes related work, and \refsec{conclusion} is dedicated to conclusions and future work.

\section{Lanes and Crowds}
\label{sec:background}

Although many definitions exist, a crowd can generally be described as ``a large group of individuals gather together in the same physical area for some duration of time''. Crowds often appear at busy public locations, such as train stations, airport terminals, football stadiums, theaters, city squares, or shopping malls. The behavior of the crowd is the results of the interactions between individuals. According to Helbing and Molnar~\cite{helbing1995social}, these interactions are local: each individuals influences only the people nearby. Describing a crowd using only local information is thus a natural representation. 

Martella et al.~\cite{martella2014crowd} proposed the idea of representing the \emph{texture} of crowds using \emph{proximity graphs}. Formally, a proximity graph is a form of spatio-temporal graph where nodes represent individuals. Time is discretized into fixed-sized timesteps and two nodes are connected by an edge at a timestep if these two individuals have been within physical proximity of each other during that time, i.e., their distance has been less than some predetermined distance. Note that proximity graphs do not store any absolute localization data, they only describe the local ``view'' of each individual. Furthermore, we assume edges do not store any information on the physical distance between nodes. Previous work~\cite{martella2014proximity} focused on methods for extracting proximity graphs from real-world noisy data obtained using proximity sensors.

\begin{figure}
  \centering
  \includegraphics[width=0.45\textwidth]{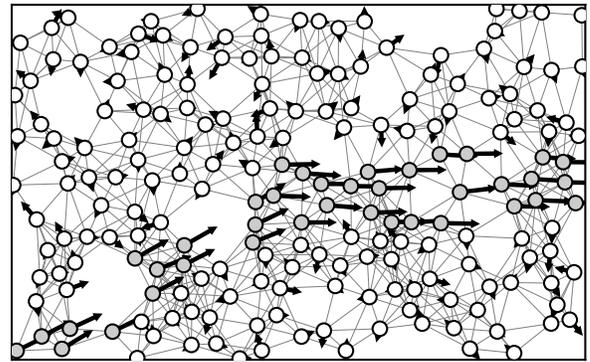}
  \caption{Example of lane in a crowd from top-down view.}
  \label{fig:example_of_lane}
\end{figure}

\reffig{example_of_lane} shows an artificial example of the proximity graph for a crowd. Points represents individuals and arrows indicate their direction and speed of movement. This particular example show non-random behavior: a lane has emerge since nodes are flowing from the bottom-left corner to the right-hand edge. According to Helbing et al.~\cite{helbing2001self}, the formation of lanes in crowds is a naturally occurrence. Individuals moving towards a target navigate the environment according to their own personal preferences. However, while moving through a dense crowd, they often need to step aside to prevent collisions with others. To minimize these interactions, it is beneficial for the walkers to follow behind someone moving in the same direction. The result of this local behavior is stable ``highway''-like lanes through the crowd.

One quick glance is sufficient to recognize that the highlighted nodes in \reffig{example_of_lane} have formed a lane. However, this observation is informal and relies on the intuition of the observer. Based on this intuition, we can define three criteria for a lane.

\begin{itemize}
\item[\textbf{(R1)}] Members of a lane move in a \textbf{similar direction} and have a \textbf{similar speed}. However, since each individual only influences its local neighborhood, each lane member should have a movement vector similar only to surrounding members. The lane as a whole can have curves and movement speed is not uniform.  Though these changes are gradual and do not happen abruptly.  Lanes move similar to how a river flows through a landscape.

\item[\textbf{(R2)}] A lane must be \textbf{connected}. In other words, if one were to create a link between each pair of lane members that are in proximity of each other, the result must be one connected unit. A lane never consist of multiple disjoint segments.

\item[\textbf{(R3)}] A lane is defined by its \textbf{border}, not by its contents. We identify lanes due to the abrupt transition between the movement inside and outside the lane. However, this border might not always be well-defined and can be ambiguous. This happens, for example, when someone leaves or joins the lane, thus blurring the line between the lane and the crowd.
\end{itemize}

\section{Algorithm for Lane Detection}
\label{sec:algorithm}
Our goal is to design an algorithm that extracts lanes from proximity graphs. First, we discuss the challenges of designing such an algorithm. Second, we present our lane-detection solution.

\subsection{Challenges}
The input of our lane detection algorithm is a proximity graph with nodes $\{v_1, \ldots, v_n\}$ and edges $E$ where $(v_i, v_j, t) \in E$ indicates that nodes $n_i$ and $n_j$ were close to each other at timestep $t$. The output should be, for each timestep $t$, the lanes detected at that moment in time. An important decision is how to deal with nodes that are not part of a lane, such as isolated nodes or stationary crowds. We have chosen to assign each of these groups to their own cluster. This simplifies the problem of lane detection into a unsupervised classification problem where the goal is to partition the nodes into ``coherent'' clusters for every moment in time. Each cluster consists of people showing similar behavior.

\subsubsection{Analysis of Proximity Graph}

Our initial attempts at lane detection were built on the following premise: choose time window $W$, aggregate data for every $W$ consecutive timesteps into a single dataset, and partition the resulting graph. However, this showed poor results since graph partitioning algorithms rely heavily on the presence of high density within each cluster. Proximity graphs have spatial nature in their topology, resulting in low intra-cluster density. In our experience, off-the-shelf graph partitioning algorithms and community detection algorithms tend to split long lanes into several separate clusters. 

The fundamental problem is that the definition of lanes roots deeply in the notion of ``distance'' and ``velocity'', which are difficult to formalize for proximity graphs. To be able to define these terms more explicitly, we embed the nodes into a two-dimensional space. An important observation is that the location determine by such embedding does not need to be highly accurate. For lane detection, only the local neighborhood of each node is relevant, so it is sufficient if the position of each node is accurate relative only to the nodes that surround it.

We found that techniques from graph drawing are suitable to calculate the embedding. Since proximity graphs change over time, the embedding is repeated for each timestep to adapt the previous embedding to the new topology. This adaptation produces movements of the nodes over time. The nodes can be clustered based on their position and velocity.

\subsubsection{Selection of Clustering Algorithm} 

Choosing the right clustering method is non-trivial since lane detection present a trade-off between two problems: \emph{transitivity} and \emph{ambiguity}. 

On the one hand, lanes can be of any arbitrary shape and they are often elongated. This means many nodes within a lane are only indirectly connected to each other. If there is a strong relation between nodes $a$ and $b$ and between $b$ and $c$, then the nodes $a$, $b$, and $c$ all belong to the same cluster, even if the relation between $a$ and $c$ is weak. Transitivity must be taken into account.

On the other hand, real-world crowds often act chaotic and clustering based on individual links between nodes is sensitive to noise. For example, consider the scenario where a person leaves the lane and joins the stationary crowd. During this transition, this person will have both a strong relation with the lane and the crowd. The clustering method should correctly interpret these ambiguous links:  a single``bridge'' between two clusters should be not be sufficient evidence that the clusters should be merged.

Centroid-based clustering methods, such as $k$-means~\cite{hartigan1979algorithm}, Mean shift~\cite{comaniciu2002mean}, or EM~\cite{moon1996expectation}, are not suitable for lane detection due to transitivity. Lanes lack a ``center'', making detection of elongated lanes impossible. Hierarchical methods, such as SLINK~\cite{sibson1973slink}, are unfit due to ambiguity since a single ``noisy'' link can cause a lane to be undetectable.

The clustering method that respect both aspects of lane detection is \emph{density-based clustering}. This class of algorithms is built on the idea that clusters correspond to dense groups of points that are separated by sparse regions. These algorithms detect clusters of any arbitrary shape, thus incorporating transitivity. They also deal well with noise, since a few outliers do not yield sufficiently high density. High quality results were obtained using DBSCAN~\cite{ester1996density}.

\vspace{-0.1cm}
\subsection{Algorithm Description}
Our lane-detection method consists of two stages: {\it graph embedding} and {\it density-based clustering}.


{\bf Graph embedding.} For the first stage, we embed the nodes into two-dimensional space using the traditional force-directed algorithm by Fruchterman and Reingold~\cite{fruchterman1991graph}. Force-directed graph embedding is a well-studied topic and many algorithms exist~\cite{kobourov2012spring}, but all follow a similar approach. Forces are assigned among pairs of vertices: attractive force between pairs connected by an edge and repulsive force between remaining pairs. The behavior of the system is simulated until an equilibrium state is reached. 


In our method, nodes are initially randomly placed and forces are simulated until equilibrium is reached. For subsequent timesteps, we use the resulting positions from the previous run as initial positions for the next run. This allows for incremental update of the node's positions and results in movement of the nodes over time. Computational cost is low since few iterations are needed per timestep to reach equilibrium.

{\bf Density-based clustering.} Next, we cluster the nodes using DBSCAN~\cite{ester1996density}, since it has proven to provide high-quality results~\cite{ester1998clustering} and scales to large datasets~\cite{zhou2000approaches}. DBSCAN takes two parameters: a radius value $\varepsilon$ and the minimum number of points $MinPts$ that should lay within this radius. More specifically, let $d_{ij}(t)$ measure the ``similarity'' between nodes $v_i$ and $v_j$ at time $t$. Clearly, $d_{ij}(t)$ can be defined in many different ways. We discuss several options for  $d_{ij}(t)$ in \refsec{evaluation}.
   The $\varepsilon$-\emph{neighborhood} of a node $v_i$ at time $t$ is the set of all nodes $v_j$ such that $d_{ij}(t) < \varepsilon$. 
  
A node is referred to as a \emph{core node} if the size of its $\varepsilon$-neighborhood is at least $MinPts$. Intuitively, core nodes are all data points ``near the core'' of the cluster since they have many neighbors in their proximity. Non-core nodes are found at the ``border'' of a cluster.

DBSCAN starts at an arbitrary node $v$. If $v$ is a not a core node, it is labeled as noise and the procedure repeats at the next unlabeled node. If $v$ is a core node, a new cluster $C$ is created containing node $v$. The cluster is now iteratively expanded by repeatedly adding every unlabeled node which is within $\varepsilon$ distance of any core node already in $C$. Once the cluster is complete, the entire procedure is repeated for the next unlabeled node.

There are different ways to handle noise points afterwards. We have chosen to assign each noise node to its own singleton cluster. Note that DBSCAN is not deterministic, non-core nodes can be assigned different clusters depending on the order in which nodes are processed. We randomize the processing order for each run.

\section{Simulation Model}
\label{sec:model}

To evaluate the quality of our lane detection algorithm, we require a simulation model which accurately models lanes in a crowd. Many models for crowd simulations have been proposed, most notably the social force model~\cite{helbing1995social} and its many variations (see survey by Castellano et. al.~\cite{castellano2009statistical}). However, while these models simulate realistic crowd dynamics, the behavior that emerges is not controlled. For example, the social force model~\cite{helbing1995social} shows lane formation, but these lanes form organically and are not planned. To evaluate the accuracy of our lane detection method, our simulations need lanes to form according to some given ground truth. To the best of our knowledge, no such model currently exists.

\begin{figure}
  \centering
  \includegraphics[width=0.35\textwidth]{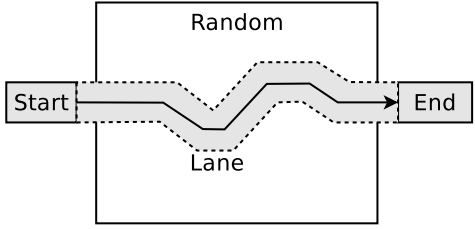}
  \caption{Example of lane going through the crowd.}
  \label{fig:regions}
\end{figure}

We propose a simple probabilistic model that exhibits controlled lane formation. Our model is based on random walks on the two-dimensional grid, i.e., each walker has integer coordinates. Initially, walkers are randomly placed in certain areas. Time passes in discrete steps. During each timestep, each walker can take one step in one of the four cardinal directions (north, east, south, west) according to predefined behavior. There are two types of behavior: \emph{random walkers} and \emph{lane walkers}.

\textbf{Random walkers} model a nearly stationary crowd. We define a rectangular region in which random walkers are initially placed at random locations (see \reffig{regions}). This region can be seen as a top-down view of a public area (e.g., city square, train station, airport terminal). During each timestep, each random walker behaves according to the following  rules:

\begin{itemize}
 \item If outside the region, take one step back.
\item Otherwise:
\begin{itemize}
\item With probability $p$: stay at current location.
\item With probability $1-p$: take one random step.
\end{itemize}
\end{itemize}

\textbf{Lane walkers} model the lane going through the crowd. For each lane, we define a path consisting of a series of line segments (see \reffig{regions} for an example). Lane walkers are initial placed at the start of the path and they follow the path until they reach the end. During each timestep, every lane walkers adheres to the following rules:

\begin{itemize}
\item With probability $q$, follow the lane. Find point $a$ on the path closest to the position $b$ of the walker.
\begin{itemize}
\item If $\|a-b\| = w > w_\text{max}$, take step in direction of $a$.
\item Otherwise, take one step in the direction tangent to line segment $ab$, i.e., follow the direction of the lane.
\end{itemize}
\item With probability $1-q$:
\begin{itemize}
  \item With probability $p$: stay at current location.
  \item With probability $1-p$: take one random step.
\end{itemize}
\end{itemize}

\begin{figure}
  \centering
  \includegraphics[width=0.4\textwidth]{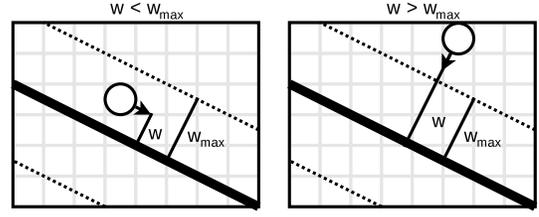}
  \caption{Two scenarios of a lane walker following a path.}
  \label{fig:scenarios}
\end{figure}

\reffig{scenarios} illustrates the walker following the lane. The parameter $w_\text{max}$ controls the maximum width of the lane. If $w > w_\text{max}$, the lane walker has wandered too far off from the lane and must move closer, thus limiting the maximum width to $2w_\text{max}$. If $w < w_\text{max}$, the lane walker must follow the direction of the lane. To keep walkers aligned on the grid, they take one horizontal step with probability $\frac{|dx|}{|dx|+|dy|}$ or one vertical step with probability $\frac{|dy|}{|dx|+|dy|}$. The average movement vector is thus $[dx~dy]^T$.

If walker $A$ wants to move to a new location which already occupied by another walker $B$, then $A$ is allowed to ``push'' $B$ by forcing it to move to one of the three remaining locations adjacent to $B$. This pushing mechanism models people stepping aside for others and is necessary to prevent bottlenecks where lane walkers are blocked by stationary random walkers. If all three adjacent locations are already occupied, the move by $A$ fails and it remains at its current location. In other words, ``pushing'' is not transmissible: walkers which get pushed cannot also push other walkers.

The parameters $p$ and $q$ control the difficulty of detecting the lane. For $p=0$, the random walkers are completely stationary and only the lane walkers move, while for $p=1$ the random walkers act chaotic. For $q=1$, the lane walkers move at maximum speed, while for $q=0$ the lane walkers show same behavior as random walkers. Changing the value of $p$ or $q$ has an impact on the difficulty of lane detection.

\ignore{
\begin{figure}
  \centering
  \includegraphics[width=0.4\textwidth]{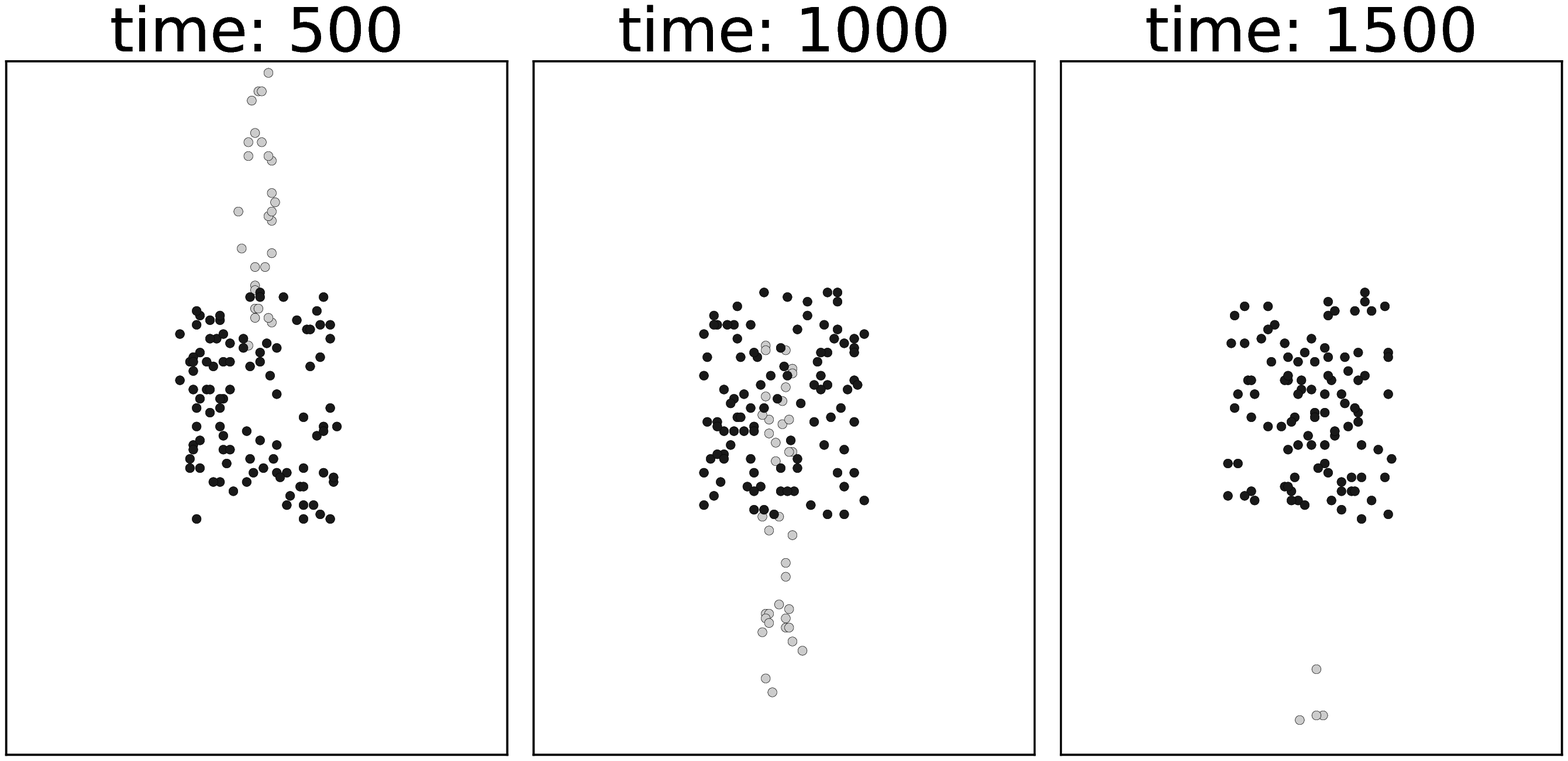}
  \caption{Example of lane simulation at different moments in time. Black and white dots correspond to random and lane walkers, respectively.}
  \label{fig:model_example}
\end{figure}

\reffig{model_example} shows an example of the simulation of a straight lane at different moments in time.
}

\section{Experimental Setup}
\label{sec:evaluation}

We evaluated our lane detection algorithm as described in \refsec{algorithm} using data generated using the model from \refsec{model}. We describe the three similarity functions and the three scenarios we consider. 

\subsection{Similarity Scores}

As discuss in \refsec{algorithm}, we are required to define a function $d_{ij}(t)$ that measures the similarity between two nodes. Low scores indicates a strong relation (i.e., nodes belonging to the same lane), while high scores indicate a weak relation. We explore three possible options for this function. The parameter $W$ is the size of the window, it determine how far we look ``back in time''.

\textbf{ Score function A}: Calculate the maximum physical distance between two nodes over the last $W$ timesteps.  The intuition is that two nodes belong to the same lane if they are physically close to each other for a long period of time. Let $p_i(t)$ be the position of node $v_i$ at time $t$. The score function is defined as follows: 

\begin{equation*}
d_{ij}^A(t) =  \max_{0 \leq dt \leq {W}} \| p_i(t-dt) - p_j(t-dt) \|.
\end{equation*}

\textbf{Score function B}: One issue with option A is that one might need a very large window size to detect the lane since two nodes can physically close for a longer duration of time while not belonging to the same lane (for example, with a horseshoe shaped lane). Alternatively, define the velocity vector $s_i$ of node $v_i$ as the average distance traveled per timestep over the last $W$ timesteps:

\begin{equation*}
s_i(t) = \frac{p_i(t) - p_i(t - W)}{W}.
\end{equation*}

Given the current position and velocity of a node, we can predict its expected position $T$ timesteps into the future. 

\begin{equation*}
d_{ij}^B(t) = \max \left[ \| p_i(t) - p_j(t) \|,\| (p_i(t) + Ts_i(t)) - (p_j(t) + Ts_j(t)) \| \right].
\end{equation*}

\textbf{Score function C}: Instead of comparing the expected future position of two nodes, we can also compare only the expected \emph{displacement}. The intuition is that if two nodes are close to each other and show similar displacement, they most likely belong to the same lane. A simple way to formalize this is as follows: 

\begin{equation}
d_{ij}^C(t) = \max \left[ \| p_i(t) - p_j(t) \|,  T~\| s_i(t) - s_j(t) \| \right].
\end{equation}

\begin{figure}
  \includegraphics[width=0.4\textwidth]{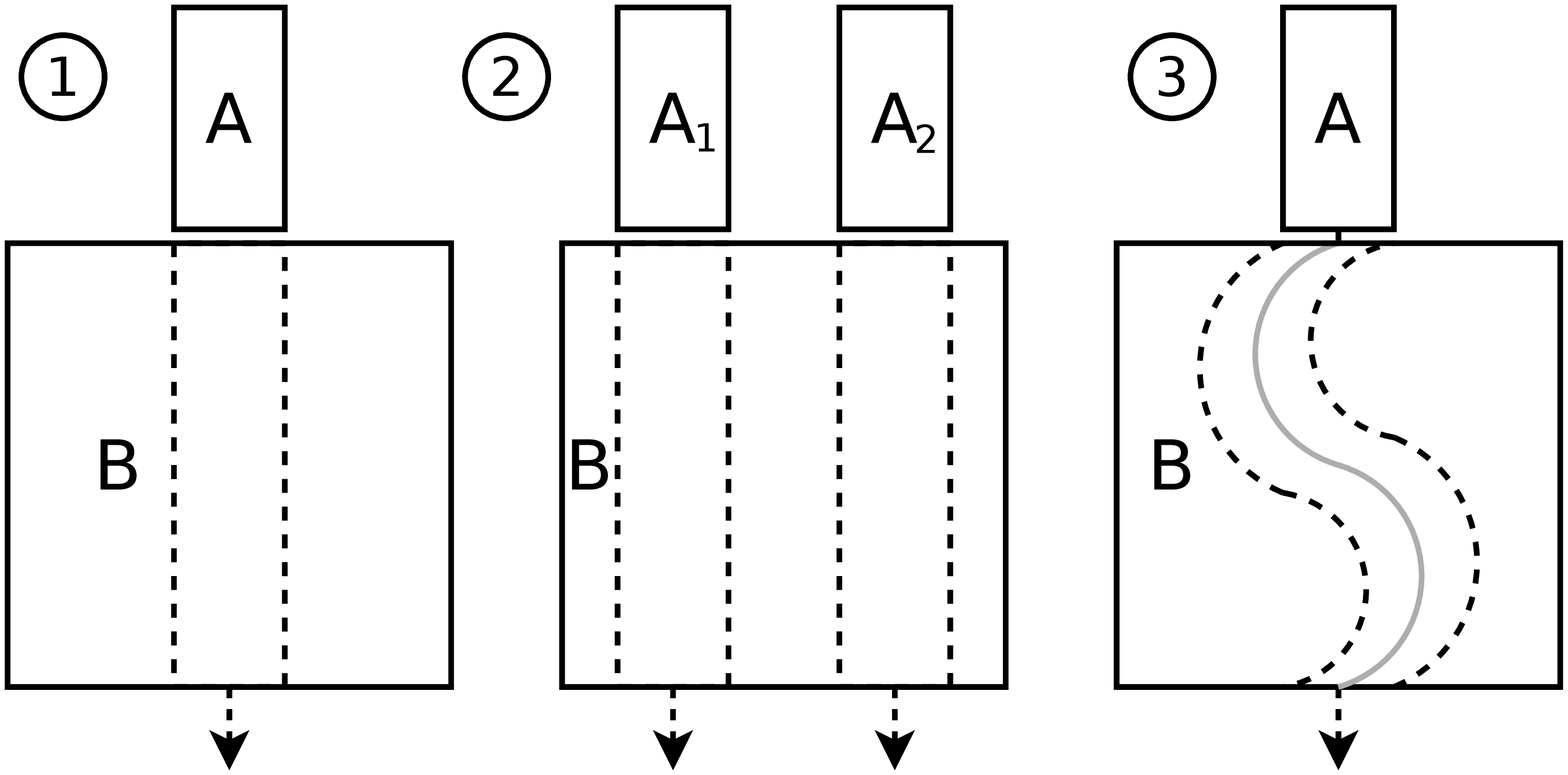}
  \caption{Three different lane scenarios used: (1) one straight lane, (2) curved lane, (3) two parallel straight lanes.}
  \label{fig:diagrams}
\end{figure}

\subsection{Scenarios}

For evaluation, we consider a scenario where random walkers are placed in a square region of $100 \times 100$ units (see \reffig{diagrams}). Lane walkers are placed in regions north of this square and walk south. The lane regions have width $w$ and height  $100 \times 100 / w$. The density of both regions is equal to ensure the number of random and lane walkers is equal. Unless noted otherwise, we set $w=10$, $p=0.2$, $q=0.5$ and density is $0.3$. We further experiment with these parameters in \refsec{results_sec}.

Our algorithm is performed during each timestep, starting at time $W$ and ending either once the last walker exits the region or until $1000$ timesteps have passed. For every timestep $t$ of the simulation, our algorithm yields a partition $X(t)=\{X_1(t), \dots, X_n(t)\}$ of the population into cohesive clusters.
The ground-truth clusters of the model are $\{R, L\}$ where $R$ is the set of random walkers and $L$ is the set of lane walkers. For the scenario with two lanes, there are three ground-truth clusters. We use the normalized mutual information~\cite{vinh2010information} (NMI) score to measure the correlation between the two partitions. The range is between $0$ (no correlation) and $1$ (perfect clustering). The reported NMI is the average over the entire simulation.

\section{Empirical Evaluation}
\label{sec:results_sec}
In this section we present the preliminary results of our method. In \refsec{results_tuning}, we evaluate the three proposed similarity functions and tune the parameters of the algorithm for a simple scenario. In \refsec{results_scenarios}, we consider a variety of scenarios with lanes of different widths and shapes. In \refsec{results_resilience}, we evaluate the resilience of our method by varying the parameters of the simulation model. 

In \refsec{results_tuning}, \ref{sec:results_scenarios}, and \ref{sec:results_resilience}, the graph embedding phase of the algorithm is omitted, i.e., coordinates from the simulation are directly used for clustering. This allows for evaluation of DBSCAN in isolation. Finally, in \refsec{results_embedding}, we revisit the problem of graph embedding.

\subsection{Method Tuning}
\label{sec:results_tuning}

\begin{figure}
  \centering
  \includegraphics[width=\columnwidth]{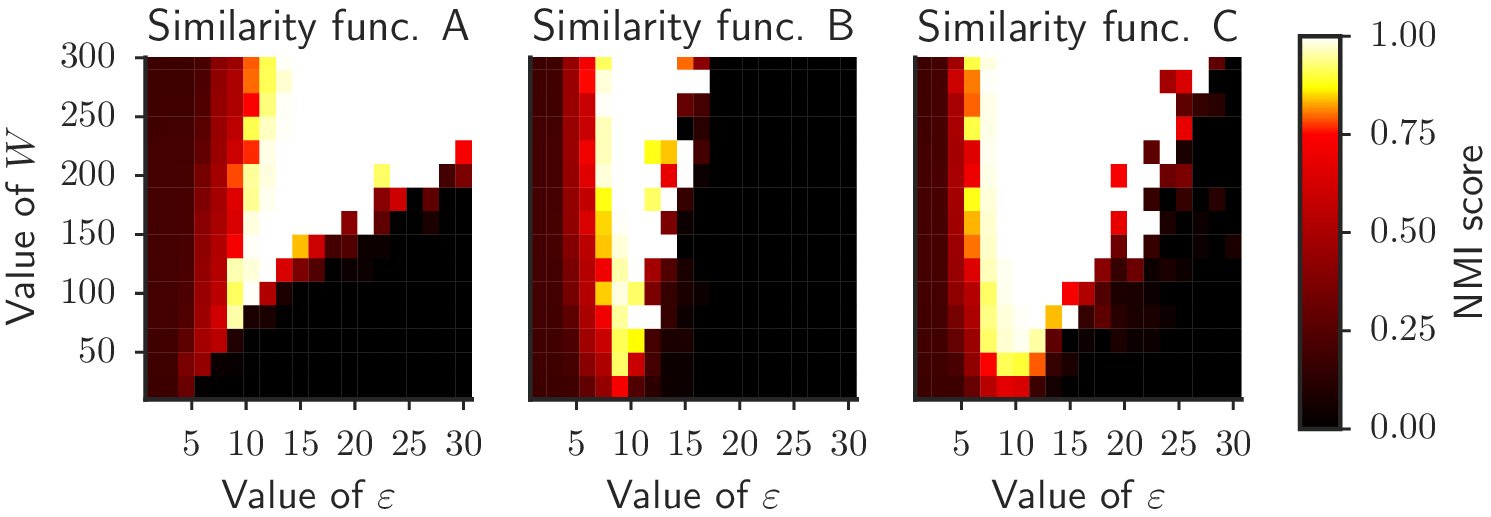}
  \caption{One straight lane, different window sizes.}
  \label{fig:results_window_size}
\quad
  \centering
  \includegraphics[width=0.7\columnwidth]{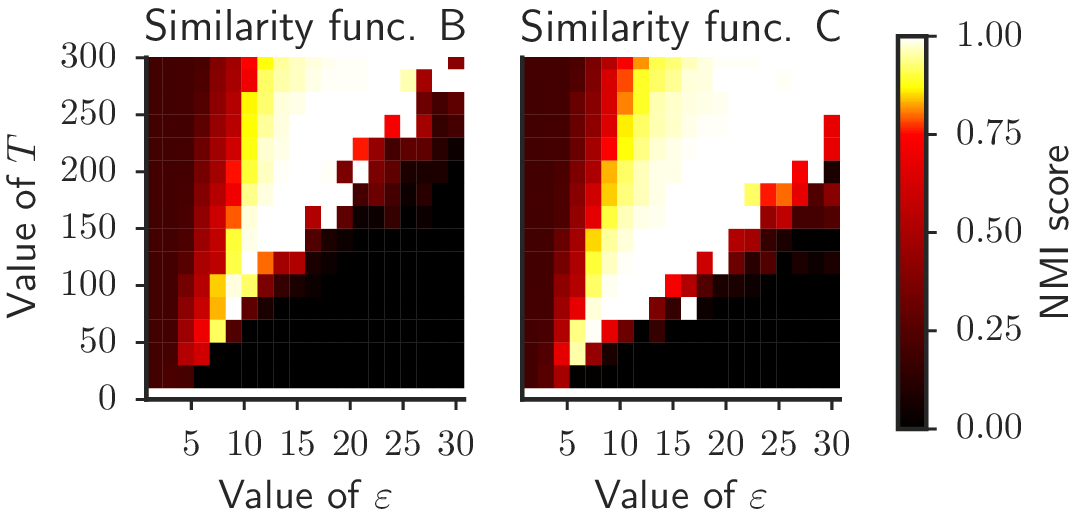}
  \caption{One straight lane, different values of $T$.}
  \label{fig:results_baseline}
\quad
  \centering
  \includegraphics[width=\columnwidth]{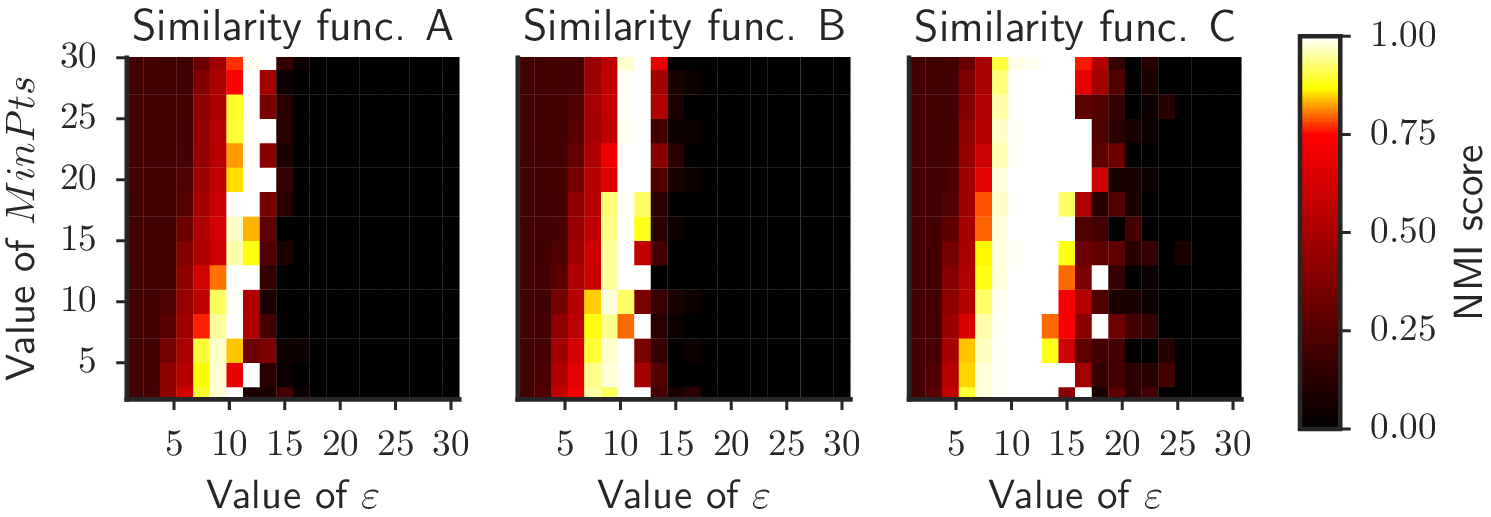}
  \caption{One straight lane, different values of $MinPts$.}
  \label{fig:results_min_pts}
\end{figure}

In this section, we focus on tuning of the parameter for DBSCAN. Four parameters are of interest: $\varepsilon$, $T$, $W$, and $MinPts$. Unless noted otherwise, we use values $T=100$, $W=100$, and $MinPts=15$. We only consider the simple scenario of one straight lane.

The crucial parameter is $\varepsilon$. For all three similarity functions, this parameter can be interpreted as the maximal physical distance that is allowed between two nodes over some period of time. If $\varepsilon$ is too small, then DBSCAN is too rigid, and many tiny clusters appear. If $\varepsilon$ is too large, then DBSCAN is too tolerant and all nodes collapse into a single cluster.

First, we consider how the window size $W$ affects the results. \reffig{results_window_size} shows the results for different window sizes. We see that the lower bound of $\varepsilon$ is approximately 10, regardless of the chosen similarity function. This can be explained based on the width of the lane. If $\varepsilon$ is less then the lane width, the random walkers on opposite sides of the lane are no longer connected since they are two far apart, causing the random walkers to be split into two groups.

The upper bound of $\varepsilon$ depends heavily on the chosen similarity function. For function A, the upper bound scales linearly with $W$. This is expected since larger $W$ implies that we look further back in time and thus the maximal distance between lane and non-lane walkers increases. For functions B and C, the upper bound also scales with $W$ but is limited to approximately $15$ for B and $25$ for C. This happens since the window size is used to calculate the velocity of nodes. For large values of $W$, the velocity converges to $(0,0)$ for random walkers and $(0,q)$ for lane walkers.

Next, we evaluate how $T$ affects the results for functions B and C, see \reffig{results_baseline}. In both cases, the lower and upper bound of $\varepsilon$ scale linearly with the value of $T$. This can be explained since for large values of $T$, the similarity score is dominated by the term $T~s_i(t)$. For larger $T$, the similarity score between neighboring lane walkers increases, which results in a wider valid range of $\varepsilon$. 

Now we turn our attention to how $MinPts$ affects quality. \reffig{results_min_pts} shows results for different values of $MinPts$ and $\varepsilon$. The figure shows that the algorithm is not sensitive to the value of $MinPts$: the upper and lower bound of $\varepsilon$ is minimally affected by its value. This parameter determines the robustness against outliers, but one straight lane contains little ambiguity.

From Figures~\ref{fig:results_window_size}, \ref{fig:results_baseline}, and \ref{fig:results_min_pts}, we conclude that function C performs the best. The valid range of $\varepsilon$ for this function is approximately between $10$ and $25$. This range scales linearly when increasing $T$, but is barely affected when varying $W$ or $MinPts$. The minimum value of $W$ should be $50$, but larger values make the algorithm less sensitive to the exact choice of $\varepsilon$. For the remainder of this work, we focus solely on function C.

\begin{figure}
  \centering
  \begin{subfigure}[b]{0.45\columnwidth}
 	 \includegraphics[width=\columnwidth]{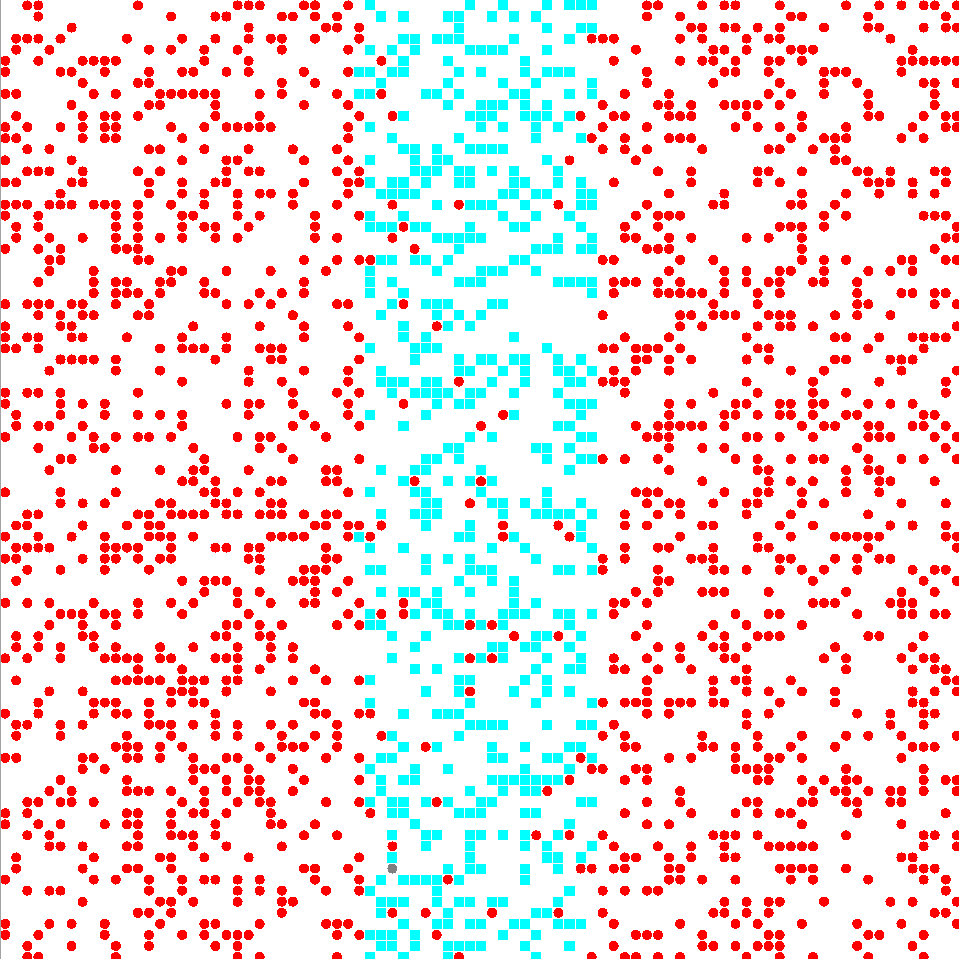}
     \caption{Lane width of 20}
  \end{subfigure}\quad %
  \begin{subfigure}[b]{0.45\columnwidth}
  	\includegraphics[width=\columnwidth]{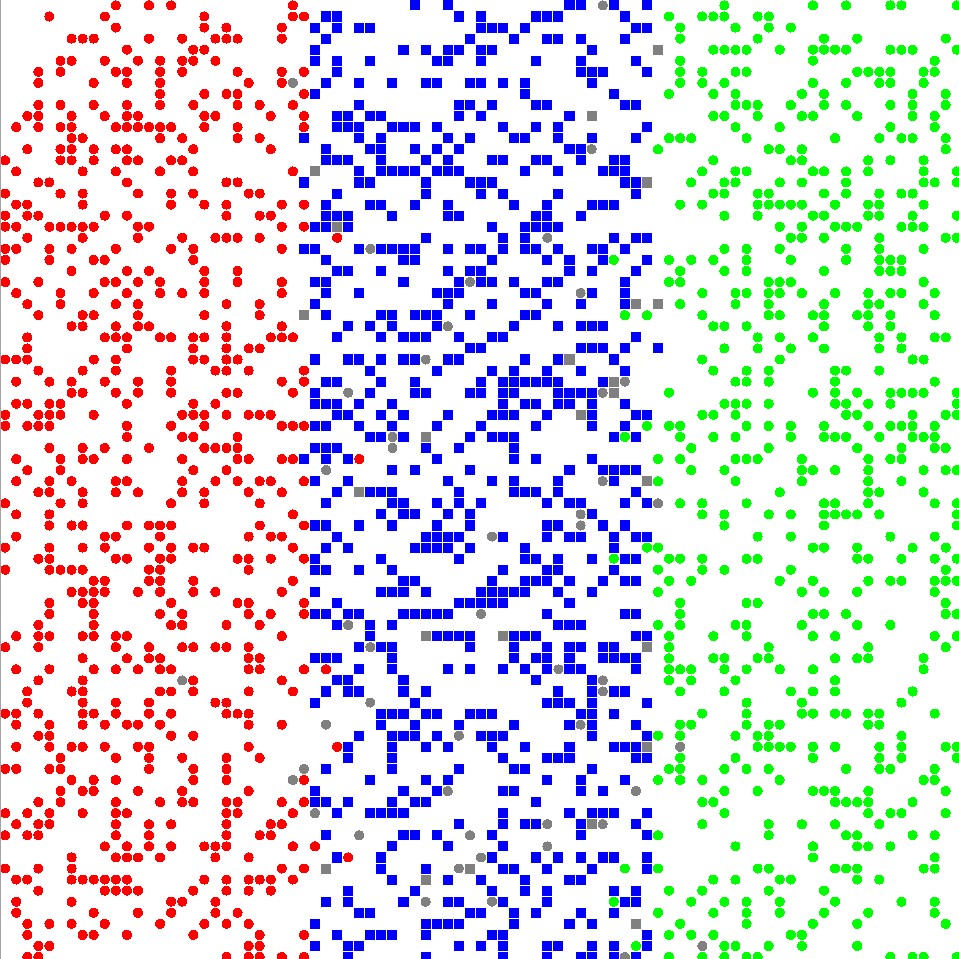}
     \caption{Lane width of 30}
  \end{subfigure}
  \caption{Lanes which are too wide cause split in stationary crowd. Each point corresponds to a walker. Different colors indicate different clusters.}
  \label{fig:vis_width_lane}
\end{figure}

\subsection{Different Types of Lanes}
\label{sec:results_scenarios}

In this section, we experiment with various types of lanes to test how well our algorithm performs in different scenarios. \reffig{diagrams} shows the three cases which we discuss.

\begin{figure}
  \centering
  \includegraphics[width=\columnwidth]{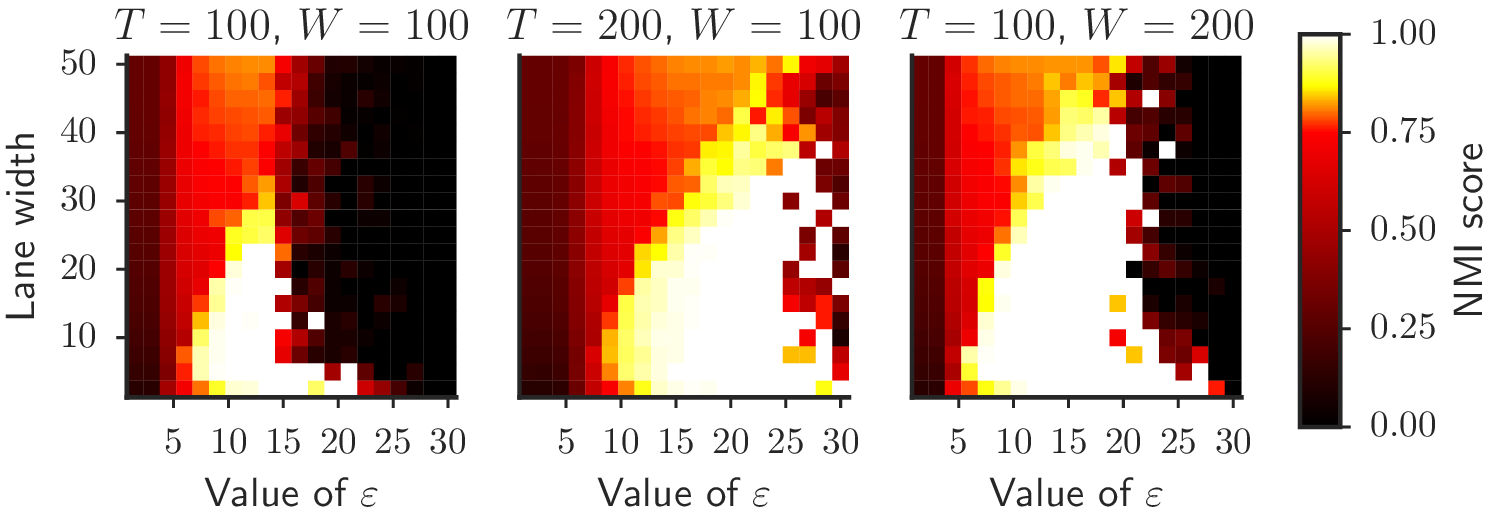}
  \caption{Baseline scenario for different lane widths.}
  \label{fig:results_lane_width}
\quad
  \centering
  \includegraphics[width=\columnwidth]{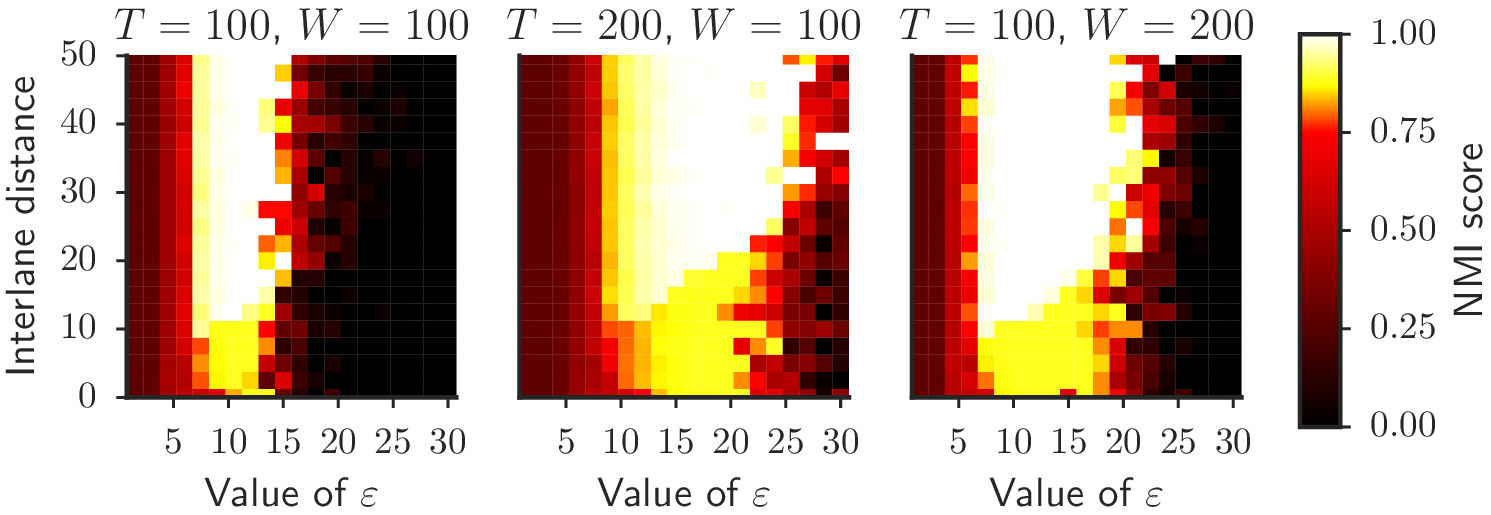}
  \caption{Two parallel lanes for variable interlane distance.}
  \label{fig:results_double_lane}
\quad
  \centering
  \includegraphics[width=\columnwidth]{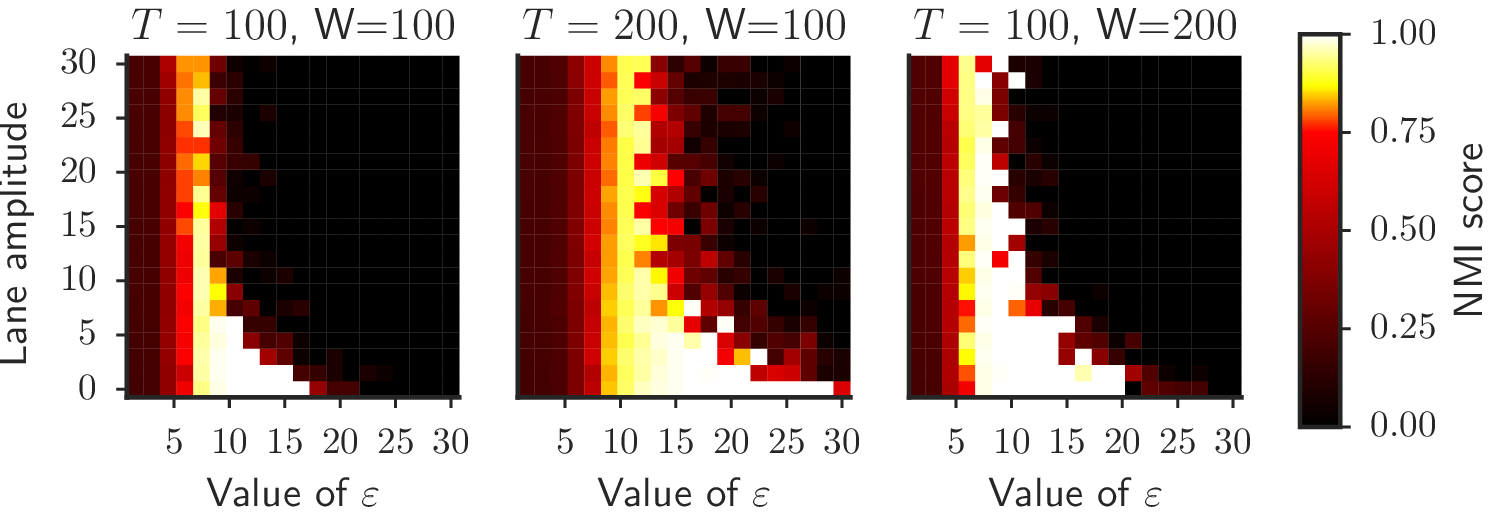}
  \caption{Sinusoidal lane for different amplitudes.}
  \label{fig:results_sinus_lane}
\end{figure}

First, we consider the scenario where the width of the lane varies between $5$ and $50$ units (i.e., the lane lane cover 5-50\% of the region of interest). \reffig{results_lane_width} shows the results. For $T=100$ and $W=100$, the lane can only be detected if its width is below approximately 20 units and $\varepsilon$ is roughly between $5$ and $15$. For wider lanes, the walkers in the stationary crowd on opposite sides of the lane are no longer considered to belong to the same cluster since they are too far apart. \reffig{vis_width_lane} demonstrates this problem. 

By increasing the value of $T$ or $W$, wider lanes can be detected. \reffig{results_lane_width} shows that both for $T=200$ and for $W=200$, lanes having a width up to 30 units can be discovered. For both cases, increasing the width of the lane also increases the lower bound of $\varepsilon$, for which the lane is detected.

Next, we consider the scenario for two parallel lanes, moving in the same direction, both of width 10, and having a fixed interlane distance. \reffig{results_double_lane} shows the results for this scenario for different interlane distances. We see that for $T=100$ and $W=100$, quality deteriorates once the lanes are less than $15$ units apart. This happens because the lanes are too close and can no longer be separated into two distinct clusters. The valid range of $\varepsilon$ is approximately in the range $10-15$.

The figure shows that doubling the value of $T$ increases the minimum separation distance to $20$. By increasing $T$, more weight is put on velocity when measuring similarity, thus making it more difficult to distinguish the two lanes. Doubling the value of $W$ does not increase the minimum separation distance and increases the upper bound of $\varepsilon$ to $20$.

Finally, to test how well our method deals with curved lanes, we consider a scenario with a single sinusoidal lane. \reffig{results_sinus_lane} shows the results for sinusoidal lanes having amplitude up to $30$ units.  Note that an amplitude of $30$ units is an extreme case, considering the height of our region is just $100$ units.

\begin{figure}
  \centering
  \begin{subfigure}[b]{0.45\columnwidth}
 	 \includegraphics[width=\columnwidth]{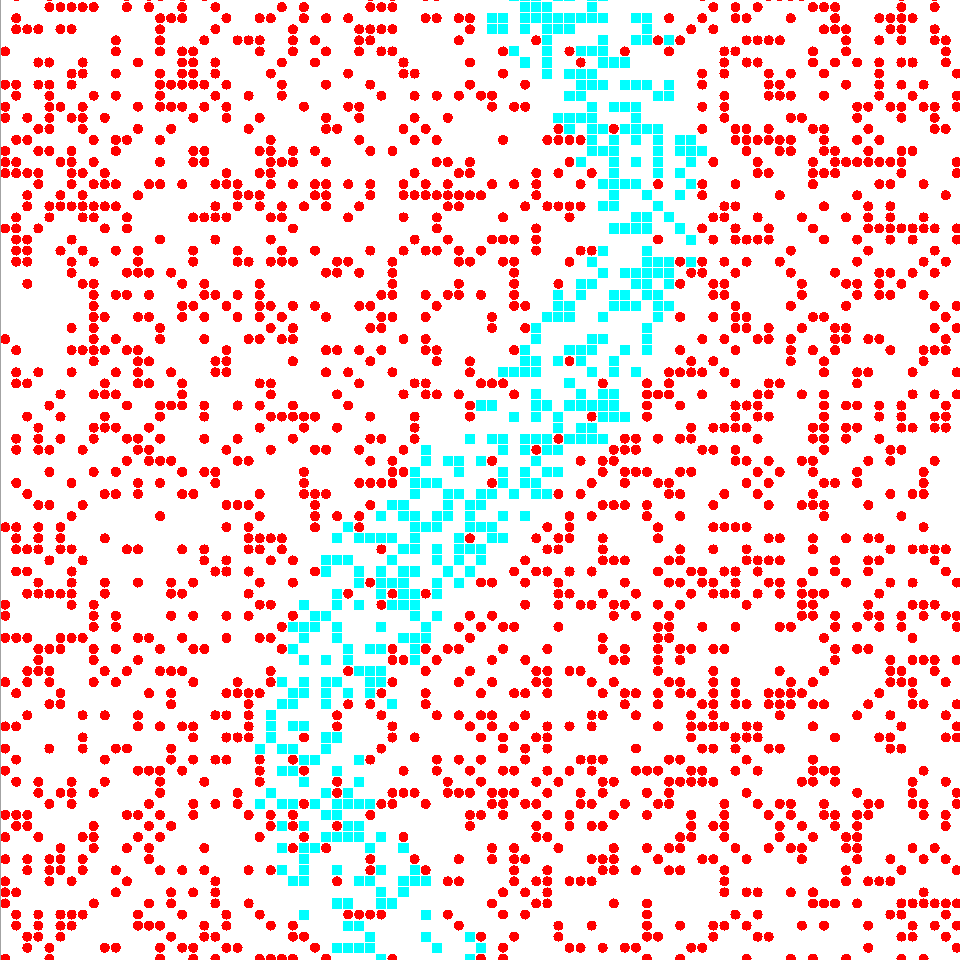}
     \caption{Amplitude of 10.}
  \end{subfigure}\quad %
  \begin{subfigure}[b]{0.45\columnwidth}
  	\includegraphics[width=\columnwidth]{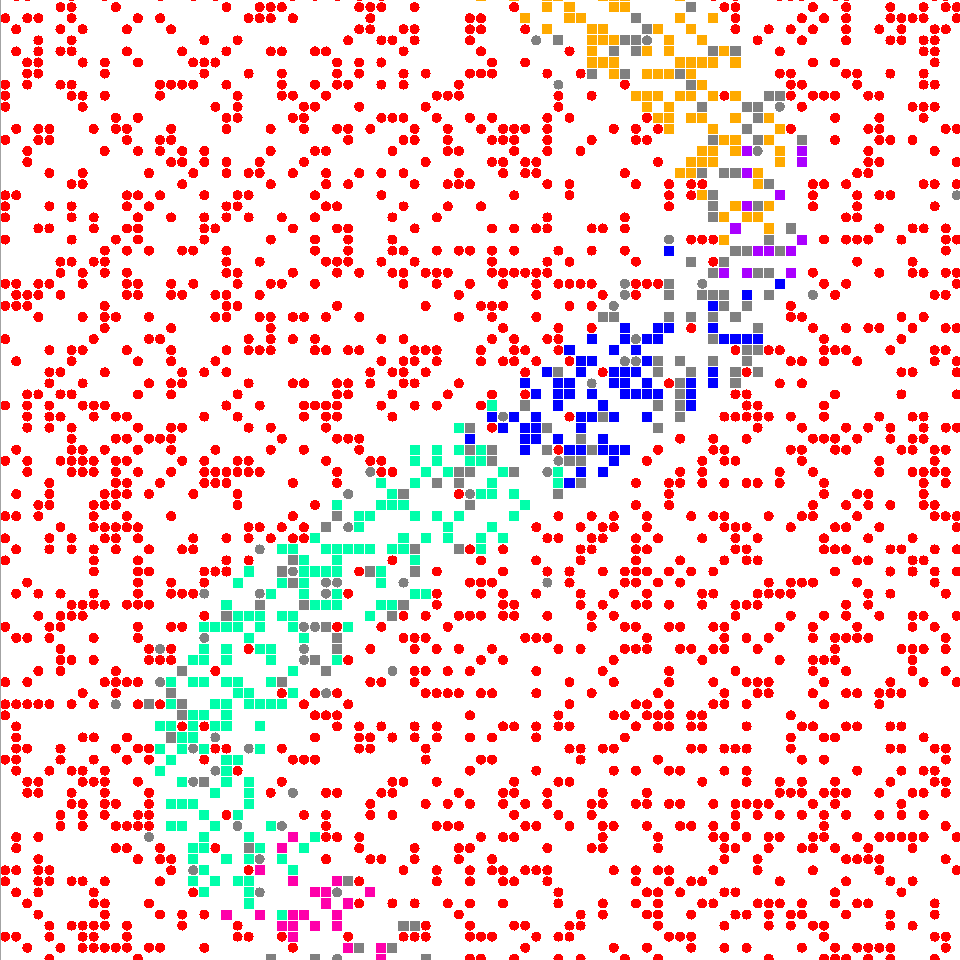}
     \caption{Amplitude of 20.}
  \end{subfigure}
  \caption{Visualization of sinusoidal lane for $T=100$, $W=100$. Each point corresponds to a walker. Different colors indicate different clusters.}
  \label{fig:vis_sinus_lane}
\end{figure}

The results show that lanes having an amplitude up to approximately $5$ units can be detected. For larger amplitudes, the algorithm tends to split the lane into multiple straight segments. \reffig{vis_sinus_lane} shows an example of this phenomenon. Increasing the value of $T$ does not change the accuracy of our method. Increasing the window size to $W=200$ significantly increases the accuracy and allows to detect lanes having an amplitude up to 30 units. The explanation is that a larger window size smooths out the sharp turning angles of the wave.

\subsection{Resilience}
\label{sec:results_resilience}

\begin{figure}
  \centering
  \includegraphics[width=\columnwidth]{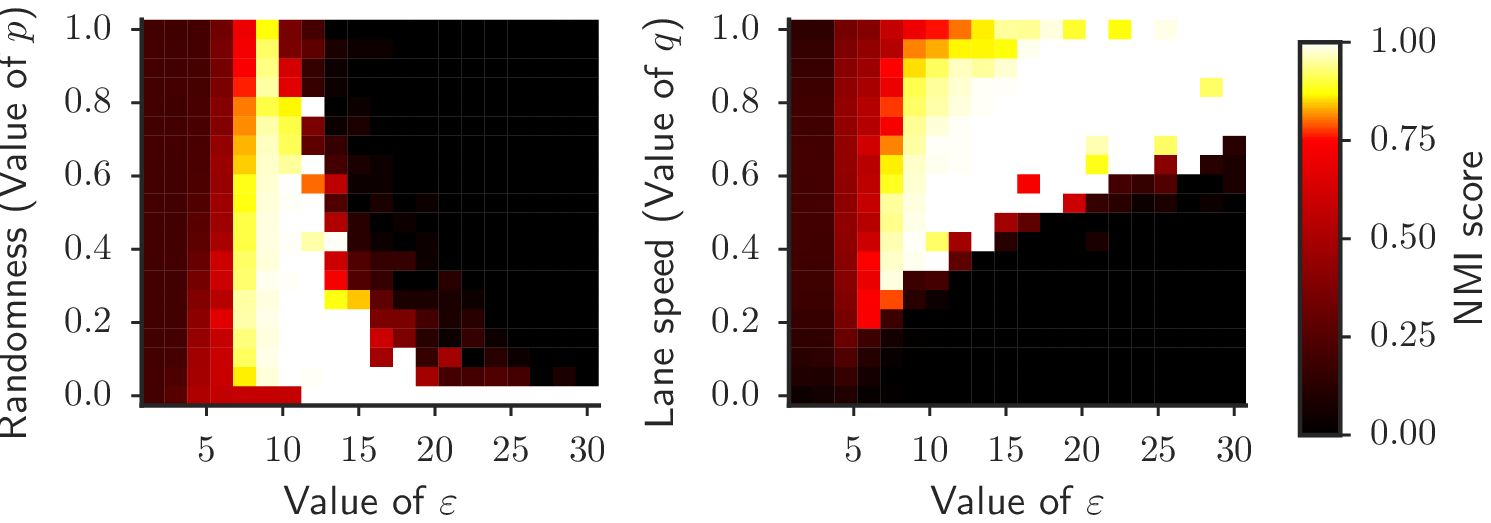}
  \caption{One straight lane for varying $p$ or $q$.}
  \label{fig:results_randomness}
\end{figure}

To test the resilience of our method, we vary  $p$ and $q$, see \reffig{results_randomness}. 

The value of $p$ determines the probability that a random walker takes a random step during a timestep. If the value of $p$ is high, detecting the lane becomes more difficult since random walkers behave erratic. \reffig{results_randomness} confirms this intuition. The lane can be detected for $p < 0.6$. 

The value of $q$ determines the probability that a lane walker follow the lane during a timestep. If the value of $q$ is too low, detecting the lane becomes difficult because the velocity of the lane is too low. The results show that the lane can only be detected if $q > 0.2$. Increasing the value of $q$ does not change the lower bound of $\varepsilon$ but its upper bound scales linear. 

For both scenarios, the detectability of the lane can be improved by using a larger window size $W$. A larger window implies that velocity is determined over longer period of time, thus containing less noise.

\subsection{Graph Embedding}
\label{sec:results_embedding}

\begin{figure}
  \centering
  \begin{subfigure}[b]{0.45\columnwidth}
 	 \includegraphics[width=\columnwidth]{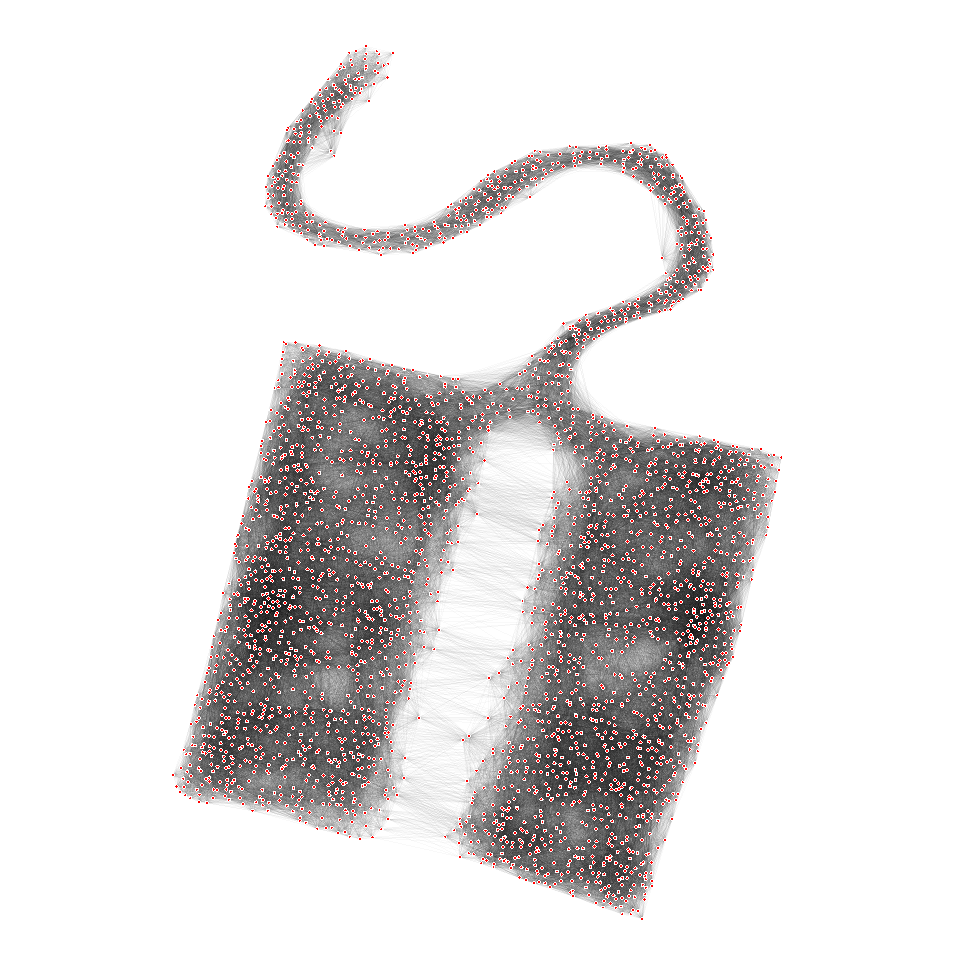}
     \caption{At timestep 0.}
  \end{subfigure}\quad %
  \begin{subfigure}[b]{0.45\columnwidth}
  	\includegraphics[width=\columnwidth]{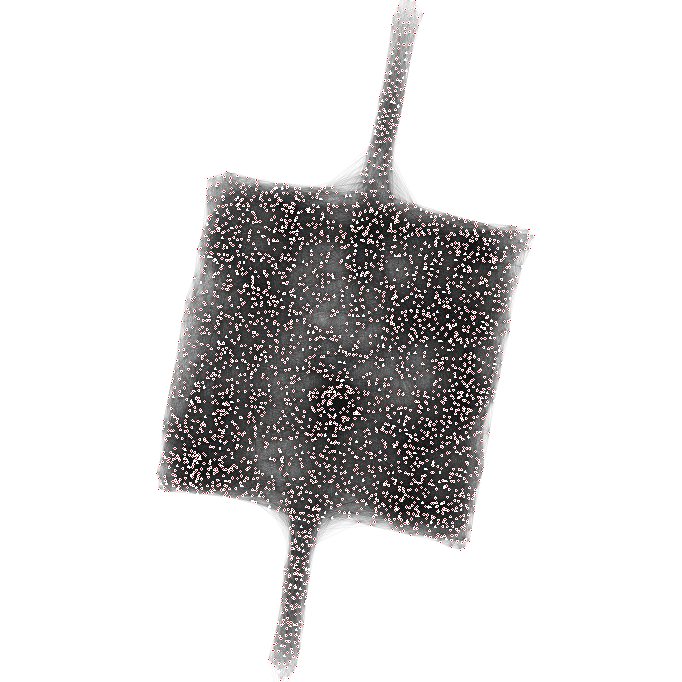}
     \caption{At timestep 300.}
  \end{subfigure}
  \caption{Results of proximity graph embedding.}
  \label{fig:vis_embedding}
\end{figure}

\begin{figure}
  \centering
  \includegraphics[width=\columnwidth]{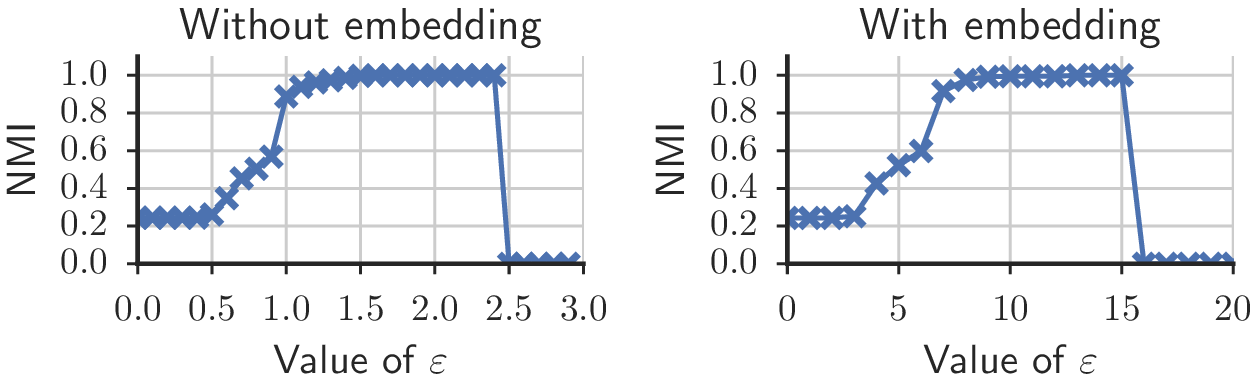}
  \caption{Accuracy with/without embedding at time. Note the different scales on the horizontal axes.}
  \label{fig:results_embedding}
\end{figure}

Up until this point, the graph embedding phase has been omitted, i.e., the absolute coordinates of the nodes are directly passed to the clustering phase. To evaluate the complete algorithm, we first generate a proximity graph and embed the nodes into two-dimensional space using force-directed embedding (see \refsec{algorithm}) before clustering the nodes. 

In practice, we find that the absolute coordinates and the coordinates found by embedding are approximately equivalent, up to scale and rotation. For example, \reffig{vis_embedding} shows an embedding of one straight lane at different moments. The proximity graph was created using a detection radius of 25 units. Force directed embedding works well for our method since the data is spatial by nature.

Since graph embedding shows excellent results, the effect on the accuracy of DBSCAN is minor. \reffig{results_embedding} shows a comparison of the obtained NMI with and without embedding for one straight lane at timestep 200. Both curves are nearly identical. Note that the difference in range of $\varepsilon$ is the result of graph embedding not preserving the scale.

\section{Related Work}
\label{sec:related_work}

Utilizing proximity graphs to analyze the behavior of people has proven to be a promising area of research. Martella et al. showed how proximity graphs can be used to mine the behavior of museum visitors~\cite{martella2016visualizing}, track people in a six-story building using only a handful of anchor points~\cite{martella2017exploiting}, and capture the social interactions at an IT conference~\cite{martella2014proximity}. However, further research on proximity graphs has been scarce.

In computer vision, the analysis of crowd behavior is an active field of research. Most work focuses on automated analysis of surveillance camera footage. We discuss some of the recent major contributions in this section. We refer to the survey by Li et al.~\cite{li2015crowded} for a comprehensive overview of research on crowd analysis from the area of computer vision.

One particular topic from computer vision which is related to lane detection is \emph{crowd behavior analysis}~\cite{li2015crowded}. These algorithms classify the behavior of people in crowds. 

For example, Rodriguez et al.~\cite{rodriguez2011data} proposed a data-driving crowd analysis approach. The algorithm works by first learning common crowd motion patterns from a large database containing crowd videos. To analyze a new video, the frames are split up into blocks which are matched to learned patches from the database. By labeling the learned patches, one can classify the behavior in different regions of the video. The authors argue that, while the number of all possible videos is infinite, the space of recognizable crowd patterns might not be all that large.

Benabbas et al.~\cite{benabbas2010motion} presented a method that can detect six crowd-related events in videos: walking, running, splitting, dispersion, and evacuation. The method works by tracking objects of interest using optical flow techniques. Next, the camera view is divided into fixed-sized blocks. For each block, the $K$ most dominant movement vectors are determined, where $K$ is a user-defined parameter. Blocks are clustered using a region-based segmentation algorithm. Finally, each cluster is classified into one of six events based on the average movement vector within the cluster.

Solmaz et al.~\cite{solmaz2012} showed how five types of behavior can be extracted from video: bottlenecks, fountainheads, lanes, arches, and blocking. Their method moves particles according to the optical flow of the video. Each region is then classified using the Jacobian matrix based on the linear approximation of the trajectories within the region. The eigenvalues of this matrix determine which of the five types the behavior belongs to.

Another topic related to lane detection and which has receive much attention in computer vision is \emph{crowd motion segmentation}~\cite{li2015crowded}. These algorithms segment the video into \emph{motion patterns}, i.e., spatial regions that have a high degree of similarity in terms of speed and direction. 

For instance, Ali et al.~\cite{ali2007lagrangian} used techniques from computational fluid dynamics for motion segmentation. A flow field is generated from frames of a moving crowd. From the flow field, a finite-time Lyapunov exponent field is constructed, which shows the Lagrangian Coherent Structures (LCS) in the underlying flow. The LCS highlight the boundaries of a flow segments and they are used for segmentation.

Kang \& Wang~\cite{kang2014fully} demonstrated how neural networks can be used to for crowd segmentation. First, they show how to use fully convolutional  neural networks to segment individuals from single static frames from videos of crowds. Next, they extend this method by integrating motion cues to capture movement, helping to separate stationary and moving crowds, and structure cues, such as walls and floors. The results show tight segmentation contours around individuals.

Zhao \& Medioni~\cite{gong2012robust} presented a method based on manifold learning and \emph{tracklets}. A tracklet is a short fragment of an object's trajectory obtained by tracking the object for short amount of time. The tracklet points are mapped to points in $(x, y, \theta)$ space, where $(x,y)$ corresponds to the image space and $\theta$ represents the motion direction in degrees between $0$ and $360$. In this 3D space, points form manifold structures each corresponding to a motion patterns. The author propose a robust manifold grouping algorithm based on Tensor Voting to extract the manifolds.


The use of proximity graphs  show two clear advantages over utilizing cameras. First, proximity graphs can provide a holistic view over a large areas. They can be used to monitor the behavior of crowds within one single build building, a small neighborhood, or even an entire city. Cameras are inherently limited to one perspective and there seems little research on how to ``join'' the image analysis from multiple cameras. Second, many computer vision techniques take a coarse-grained approach and classify regions within the image, meaning any information about individuals is lost. Our approach classifies nodes of the proximity graphs, thus retaining this fine-grained information.

Overall, we believe our method is the first lane detection algorithm designed for proximity graphs.

\section{Conclusions \& Future Work}
\label{sec:conclusion}

In this work, we present a method to detect lanes in proximity graphs. Our method combines graph embedding with density-based clustering. For evaluation, we have explored three different score functions to measure similarity between nodes. Best performance was obtained by measuring similarity as the maximum over two terms: difference in position (distance) and difference in velocity. The results show that our method can detect different types of lanes (thick lanes, parallel lanes, and curved lanes). Graph embedding performs excellent, although its computational cost is high. For DBSCAN, exact tuning of the parameters is important. Most notably, DBSCAN shows sensitivity to the value of $\varepsilon$.

For future work, we are exploring methods to automatically determine the best parameters for DBSCAN. Furthermore, we are looking into more complex scenarios. For example, opposing lanes, lanes crossing at an intersection, and lanes moving through a narrow doorway. We are also extending our simulation model to support more situations, such as people joining/leaving the lane or a lane dissolving into the crowd. Finally, we are working on obtaining real-world measurements to evaluate our method on non-synthetic datasets.

Overall, we view our work as a first step towards rich pattern recognition in proximity graphs. One can think of many types of crowd behavior identification, such as detection of congestion, social cliques, evacuations, and anomalies. Our goal is to utilize proximity graphs as a tool to enable these types of analysis.

\bibliographystyle{ACM-Reference-Format}
\bibliography{ref}

\end{document}